\documentclass{ecai}  % use option [doubleblind] for double blind submission and hiding the authors section

\usepackage{cite}
\usepackage{amsmath,amssymb,amsfonts}
\usepackage{algorithmic}
\usepackage{graphicx}
\usepackage{textcomp}
\usepackage{xcolor}
\usepackage{subfigure}
\usepackage{multirow}
\usepackage{booktabs}

\usepackage{dcolumn}
\usepackage{subfigure}
\usepackage{multirow}
\usepackage{makecell}
\usepackage{enumitem}
\usepackage{fancyhdr}

\begin{document}
	
	\begin{frontmatter}
		
		\title{Self-supervised Learning and Graph Classification
			under Heterophily}

\author[A]{\fnms{Yilin }~\snm{Ding}}
\author[A]{\fnms{Zhen}~\snm{Liu}\thanks{Corresponding Author. Email: zhliu@bjtu.edu.cn.}}
\author[A]{\fnms{ Hao}~\snm{Hao}} % use of \orcid{} is optional

\address[A]{Beijing Jiaotong University}

		\begin{abstract}
	Self-supervised learning has shown its promising capability in graph representation learning in recent work. Most existing pre-training strategies usually choose the popular Graph neural networks (GNNs), which can be seen as a special form of low-pass filter, fail to effectively capture heterophily. In this paper, we first present an experimental investigation exploring the performance of low-pass and high-pass filters in heterophily graph classification, where the results clearly show that high-frequency signal is important for learning heterophily graph representation. On the other hand, it is still unclear how to effectively capture the structural pattern of graphs and how to measure the capability of the self-supervised pre-training strategy in capturing graph structure. To address the problem, we first design a quantitative metric to Measure Graph Structure (MGS), which analyzes correlation between structural similarity and embedding similarity of graph pairs. Then, to enhance the graph structural information captured by self-supervised learning, we propose a novel self-supervised strategy for Pre-training GNNs based on the Metric (PGM). Extensive experiments validate our pre-training strategy achieves state-of-the-art performance for molecular property prediction and protein function prediction. In addition, we find choosing the suitable filter sometimes may be better than designing good pre-training strategies for heterophily graph classification.
		\end{abstract}
		
	\end{frontmatter}
	
	\section{Introduction}
	%The traditional page limit for ECAI long papers is {\bf 7 (six)} pages
	%in the required format. The traditional page limit for short
	%submissions is {\bf 2} pages.
	%
	%However, these page limits may change from one ECAI to
	%another. Consult the most recent Call For Papers (CFP) for the most
	%up-to-date page limits.
	
	In recent years, graph neural network (GNN) ~\cite{kipf2017semi-supervised,velickovic2018graph,hamilton2017inductive} has had a profound impact on machine learning tasks based on graph structure data, such as node classification, link prediction and graph classification. However, training effective graph neural networks usually requires large amounts of labeled data, which cannot be satisfied in many scenarios due to the high cost of labeling training data. This problem is exacerbated in scientific domains, such as chemistry and biology, where data labeling is resource- and time-intensive. Inspired by the success of self-supervised learning in Natural Language Processing (NLP) and Computer Vision (CV) communities ~\cite{xiao2020self-supervised}, recent studies start to utilize self-supervised methodology to pre-train a GNN on unlabeled data, and then transfer the learned model to downstream tasks, which has achieved substantial improvements. There are two key points: 1) How to select a suitable GNN based on the characteristics of the graph itself (such as heterophily and homophily); 2) how to design appropriate self-supervised learning to capture the graph structural information.
	
	Firstly, while many GNN models have been proposed, most of them are designed under the assumption of homophily, and are not capable of handling heterophily. Homophily is a key principle of many real-world networks, whereby most connections happen among nodes in the same class or with alike features. For example, friends are likely to have similar political beliefs or ages, and papers tend to cite papers from the same research area [23]. However, there are also settings where “opposites attract”, leading to networks with heterophily: linked nodes are likely from different classes or have dissimilar features. For instance, different amino acid types are more likely to connect in protein structures. In general, GNNs can be seen as a special form of low-pass filter\cite{zhu2021graph,pei2020geom}, which mainly retains the commonality of node features, inevitably ignores the difference, this mechanism may work well for homophily networks but not heterophily networks. Based on the discussion above, we conducted some experiments in the following section, to verify the hypothesis that the biograph dataset used in the paper is heterophily, and to explore the role of low-frequency and high-frequency signals on the heterophily graph classification performance.
	\begin{figure}[t]
		\begin{center}
			\subfigure[No Pre-train $ ({\rm MGS}=0.308) $]
			{
				\begin{minipage}[b]{.45\linewidth}
					\centerline{\includegraphics[scale=0.33]{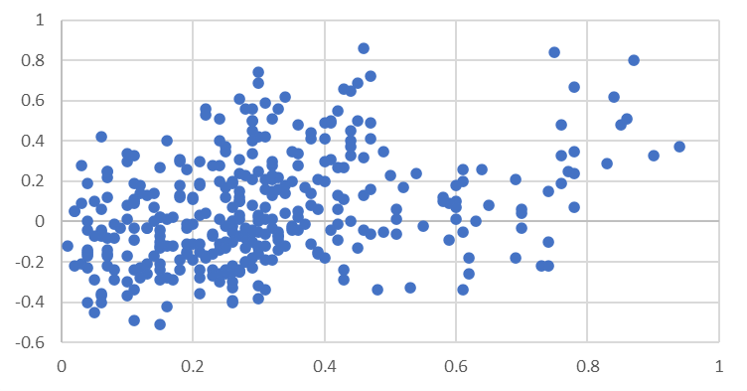}}
					\label{icml-historical}
				\end{minipage}
			}
			\subfigure[Pre-train $ ({\rm MGS}=0.700) $]
			{
				\begin{minipage}[b]{.45\linewidth}
					\centerline{\includegraphics[scale=0.33]{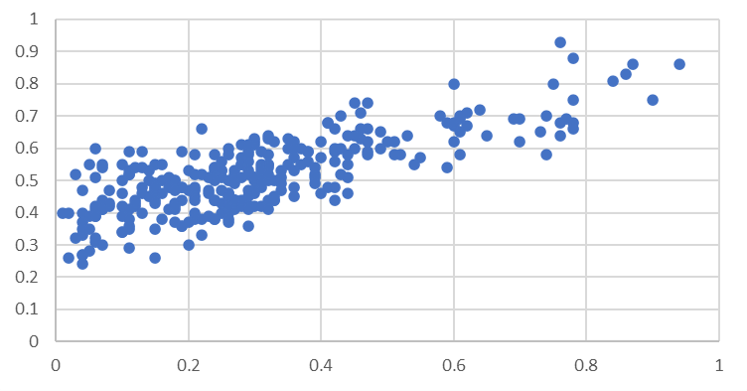}}
					\label{icml-historical}
				\end{minipage}
			}
			\label{fig1}
		\end{center}
		\caption{
			The MGS values of Graph Isomorphism
			Network (GIN) with and without pre-training on the ZINC15 dataset.
			Specifically, the horizontal and the vertical coordinates represent the Tanimoto similarity of fingerprints, and the Cosine similarity of the graph representations respectively.}
		\label{fig1}
	\end{figure} 
	
	Secondly, heterophily graph usually contains rich structural information that can be useful for graph representation learning. Previous works have made initial attempts to capture graph structure in a self-supervised manner, such as  ContextPred~\cite{hu2020strategies}, EdgePred~\cite{hamilton2017inductive}, DGI~\cite{velickovic2019deep}, PHD~\cite{li2021pairwise}, GraphLoG~\cite{xu2021self-supervised}, which achieves notable improvements on downstream tasks. However, it is still unclear how much graph structure captured by these self-supervised methods. This is a fundamental question since some pre-training methods are just proposed based on intuition, but there is no metric to measure the graph structure information captured by self-supervised pre-training strategy. 
	
	To measure the capability of pre-training strategies in capturing graph structural information, we propose a quantitative metric to Measure Graph Structure (MGS). In detail, we calculate the structural similarity and embedding similarity for each graph pair. For structural similarity, we introduce graph fingerprint, which reduces a graph to a much smaller feature set, to accurately capture the topological structure of the graph, and the Tanimoto similarity of fingerprints is selected to calculate structural similarity. For embedding similarity, we calculate the cosine similarity of the pre-trained graph embeddings. Considering these two sets of similarity values do not follow a normal distribution, we choose spearman correlation to calculate correlation between the two kinds of similarity. An example is shown in Figure. \ref{fig1}, we can observe a higher correlation between the two different similarities when the structural pre-training strategy Infomax~\cite{velickovic2019deep} is applied. Other experiments in the following sections also verify that the more structural information contained in graph embeddings, the higher MGS value is. Hence, our proposed metric MGS indeed measures the graph structure captured by pre-training strategy.
	
	Once the quantitative metric standard is identified, we further propose a novel and simple self-supervised strategy for Pre-training Graph neural networks based on the Metric (PGM). The key idea is to maximize the spearman correlation between cosine similarity of pre-trained embeddings and structural similarity of graph fingerprints. When pre-training is finished, we fine-tune the pre-trained GNN on downstream tasks. Specifically, we add linear classifiers on top of graph-level representations to predict downstream graph labels.
	
	Our contributions can be summarized as follows:
	\begin{itemize}
		\item We calculate the homophily ratio of the biological graph, and explore the performance of low-pass and high-pass filters in graph classification. Experiments demonstrate that the high-frequency signal is helpful to the graph classification. 
		\item We design a quantitative metric (MGS) to measure the capability of pre-training strategies in capturing structural information. Experiment results verify that MGS has a significantly high correlation with graph structure captured by model.
		\item We propose a novel and simple self-supervised strategy based on the metric (PGM). The strategy can pretrain GNNs at graph level and empowers the GNN model to capture graph structure.
		\item Extensive experiments demonstrate our pre-training strategies can: 1) outperform the previous baselines; 2) cooperate well with node-level strategies; 3) generalize well to different GNNs; 4) apply well to other fingerprint and similarity algorithms; 5) perform best in visualization.
	\end{itemize}
	
	\section{Related Work}
	\subsection{Heterophily \& GNN } In general, GNNs update node representations by aggregating information from neighbors, which can be seen as a special form of low-pass filter\cite{wu2019simplifying,li2019label}. Recent studies find that the low-pass filter in GNNs mainly retains the commonality of node features, which inevitably ignores the difference, leads to poor performance on heterophily graphs\cite{zhu2021graph,pei2020geom}. To address this limitation, some work\cite{zhu2020beyond,yan2022two,zhu2021graph} accordingly designs and modifies neural architectures and demonstrates outperformance over other GNN models on several heterophily graphs. Another study\cite{bo2021beyond} find that when a network exhibits heterophily, high-frequency signals perform much better than low-frequency signals, that is high-pass filter is suitable for heterophily graph. However, previous studies are especially centered around node representation learning and semi-supervised node classification, and there is almost no research on the performance of low-pass and high-pass filters on heterophily graph classification.
	\subsection{Self-supervised Learning on Graphs} As a technique of transfer learning, self-supervised learning is proposed to extract knowledge from large-scale unlabeled data by leveraging input data itself as supervision. Self-supervised learning was initially applied in the fields of CV and NLP\cite{liu2021self}, where it augments the raw data through image rotation/clipping and sentence masking. In rencent studies, SSL has been applied to GNNs, and has received considerable attention. Among which,~\cite{hu2020strategies,hamilton2017inductive,rong2020self-supervised} do masking and predicting in nodes, edges or contexts in the topology, might not capture global structural information well and result in limited performance. Graph-level strategies including PHD~\cite{li2021pairwise}, GraphCL~\cite{you2020graph}, though do improve performance, it is not clear whether structural information is actually captured by them from graph. In contrast, our proposed pre-training strategy makes the capture of structural information more explainable.
	
	\begin{table*}
		\setlength{\tabcolsep}{1mm}
		\begin{center}
			{\caption{The homophily ratio of datasets (Left: Chemistry datasets, Right: Biology dataset).}
				\label{tab11}}
			\begin{tabular}{ccccccccc|c}
				\toprule
				&BACE&BBBP&SIDER&ToxCast& HIV&Tox21&MUV&Clintox&Proteins\\
				\midrule
				\text { Hom.ratio } & 0.56 & 0.61 & 0.57 & 0.37 & 0.64 & 0.45& 0.71 & 0.65& 0.72 \\
				\bottomrule
			\end{tabular}
		\end{center}
	\end{table*}

	\begin{table}
		
		\setlength{\tabcolsep}{1mm}
		\begin{center}
			{\caption{Test ROC-AUC score on supervised molecular and protein graph classification.}
				\label{tab12}}
			\begin{tabular}{ccc}
				\toprule 
				Dataset & Chemistry & Biology \\
				\midrule 
				GCN & 68.9  & 63.2 $ \pm $ 1.0 \\
				ChebNet & \textbf{70.0}  & 67.5 $ \pm $ 0.6 \\
				FAGCN & 67.0 & \textbf{71.0 $ \pm $ 0.5} \\
				\bottomrule
			\end{tabular}
		\end{center}
	\end{table} 
	
	\section{Problem Formulation}
	\subsection{Preliminary}
	\subsubsection{Spectral Graph Convolution}
	Spectral convolutions on graphs are defined as the multiplication of a signal $x \in \mathbb{R}^N$ (a scalar for every node) with a filter
	$g_\theta=\operatorname{diag}(\theta)$ parameterized by $\theta \in \mathbb{R}^N$ in the Fourier domain:
	\begin{equation}
	g_\theta \star x=U g_\theta U^{\top} x
	\end{equation}
	where $U$ is the matrix of eigenvectors of the normalized graph Laplacian $L=I_N-D^{-\frac{1}{2}} A D^{-\frac{1}{2}}=$ $U \Lambda U^{\top}$, with a diagonal matrix of its eigenvalues $\Lambda$ and $U^{\top} x$ being the graph Fourier transform of $x$. We can understand $g_\theta$ as a function of the eigenvalues of $L$, i.e. $g_\theta(\Lambda)$. ChebNet~\cite{defferrard2016convolutional} parameterizes convolutional kernel with a polynomial expansion $g_\theta=\sum_{k=0}^{K-1} \alpha_k \Lambda^k$. GCN defines the convolutional kernel as $g_\theta=I-\Lambda$.
	\subsubsection{Spatial Graph Convolution}
	GNNs can be regarded as message passing neural networks~\cite{gilmer2017neural}. During each message-passing iteration in a GNN, a hidden embedding $ h_{u}^{(k)} $ corresponding to each node $u \in V $ is updated according to information aggregated from $ u's $ graph neighborhood $\mathcal{N}(u)$. This message-passing update process can be expressed as follows:
	\begin{equation}
	h_{u}^{(k+1)}=\operatorname{UPDATE}^{(k)}\left(h_{u}^{(k)}, \text { AGGREGATE }^{(k)}\left(\left\{h_{v}^{(k)}, \forall v \in \mathcal{N}(u)\right\}\right)\right)
	\end{equation}
	
	where $ { AGGREGATE }^{(k)} $ is the aggregation function for aggregating messages from a node’s neighborhood, and $ {UPDATE}^{(k)} $ is the update function for updating the node representation. We initialize $h_{v}^{(0)}$ by the feature vector of node v, i.e.,$h_{v}^{(0)}x_{v}$. 
	\subsubsection{Graph Representation Learning}To obtain the entire graph’s representation $ h_{G} $, the READOUT
	function pools node features from the final iteration K:
	\begin{equation}
	h_{G}=\operatorname{READOUT}\left(\left\{h_{v}^{(K)} \mid v \in G\right\}\right)
	\end{equation}
	
	\subsection{Problem Definition}
	Given a graph $ G = (V,E) $ where $ V $ represents nodes and $ E $ represents edges. For example, a molecule can be abstracted as a graph, where nodes represent atoms and edges represent chemical bonds.
	In this paper, we aim at pre-training a GNN model on a large number of unlabeled graphs under the guidance of the data itself, and then apply the pre-trained model to the downstream datasets for graph classification in a supervised manner. Specifically, we expect the pre-trained model can capture the graph structural information, which can be transferred to downstream tasks such as molecular property prediction. 
	
	\section{Graph Classification under Heterophily}
	Research on graph with homophily/heterophily has mostly focused on node classification, and there is almost no research on the performance of low-pass and high-pass filters on graph classification. Therefore, in this section, we attempt to calculate the homophily
	ratio of biology and chemistry graphs, and conduct experiments to explore the performance of low-pass and high-pass filters in graph classification task.
	\subsection{Graph Homophily}
	Given a graph $ G = {(V, E)} $ and node label vector $ y $, the homophily
	ratio is defined as the fraction of edges that connect nodes with the same labels. Formally, we have:
	\begin{equation}
	h=\frac{\left|\left\{(u, v):(u, v) \in \mathcal{E} \wedge y_u=y_v\right\}\right|}{|\mathcal{E}|}
	\end{equation}
	
	where $|\mathcal{E}|$ is the number of edges in the graph. By definition, we have $h \in[0,1]$ : graphs with $h$ closer to 1 tend to have more edges connecting nodes within the same class, or stronger homophily; on the other hand, graphs with $h$ closer to 0 have more edges connecting nodes in different classes, or a stronger heterophily. From Table \ref{tab11}, we calculate homophily ratio of chemistry dataset  and biology dataset. Most of the datasets in this paper have homophily ratio of less than 0.5, which indicates heterophily or low homophily.

	\subsection{Different filters on Graph Classification}
	We can consider the function $g_\theta(\Lambda)$ in Equation 1 as a convolution kernel. Different $g_\theta(\Lambda)$ functions can achieve different filtering effects. 
	ChebyNet defines graph convolution based on a set of combined filters, i.e.,  $g_\theta=\sum_{k=0}^{K-1} \alpha_k \Lambda^k$, which are obtained through assigning higher weights to high-frequency basic filters. For example, for a combined filter $\Lambda^k$, the weight assigned to $u_i u_i^T$ is $\lambda_i^k$, which increases with respect to $\lambda_i$. GCN only uses the first order approximation, i.e., $L$, for semi-supervised learning. Both the low-frequency and high-frequency signals are helpful for learning node representations. To make full use of them, FAGCN\cite{bo2021beyond} designs an enhanced low-pass filter
	$ F_{L}=(\varepsilon+1)I-\Lambda $ and an enhanced high-pass filter $ F_{H}=(\varepsilon+1)I+\Lambda $, and use the attention mechanism to adaptively combine the low and high frequency signals.
	
	According to the above discussion, ChebNet can be regarded as a high frequency filter, GCN as a low frequency filter, and FAGCN as a frequency adaptive filter. To verify the graph classification performance of different filters on heterophily graphs, We carried out experimental verification on supervised graph classification. The experimental results are shown in the Table \ref{tab12}.

	From Table \ref{tab12}, we find that the performance of ChebNet is better than that of GCN and FAGCN for chemistry datasets, and FAGCN is better than GCN and ChebNet for biology dataset. The reason may be: from Table \ref{tab11}, the homophily ratio of chemistry datasets is mostly within 0.5, high-frequency information may be more important, so the high-pass filter ChebNet has a better performance; the homophily ratio of biology dataset is more than 0.5, extracting high-frequency signals alone does not improve the performance, so FAGCN, which adaptively integrate the low-frequency signals and high-frequency signals, achieves the best performance. 
	
	\section{MGS: Metric for Graph Structure}
	The purpose of pre-training is to obtain general graph representations. Therefore, the quality of graph structure information contained in pre-trained embeddings can reflect the capability of pre-training strategies in capturing structural information.
	
	So how can we measure the quality of embeddings? Intuitively, two graphs which are similar have
	similar embeddings. Therefore, we generate graph fingerprint, which captures the topological structure of the graph, and calculate the correlation between structural similarity and embedding similarity as the metric to measure the quality of graph structure information contained in pre-trained embeddings.
		\begin{figure*}[ht]
		%\centering
		\subfigure[GCN  $ ({\rm MGS}=0.510) $  \qquad FCN  $ ({\rm MGS}=0.216) $]{
			\begin{minipage}[b]{.22\linewidth}
				\centering
				\includegraphics[scale=0.23]{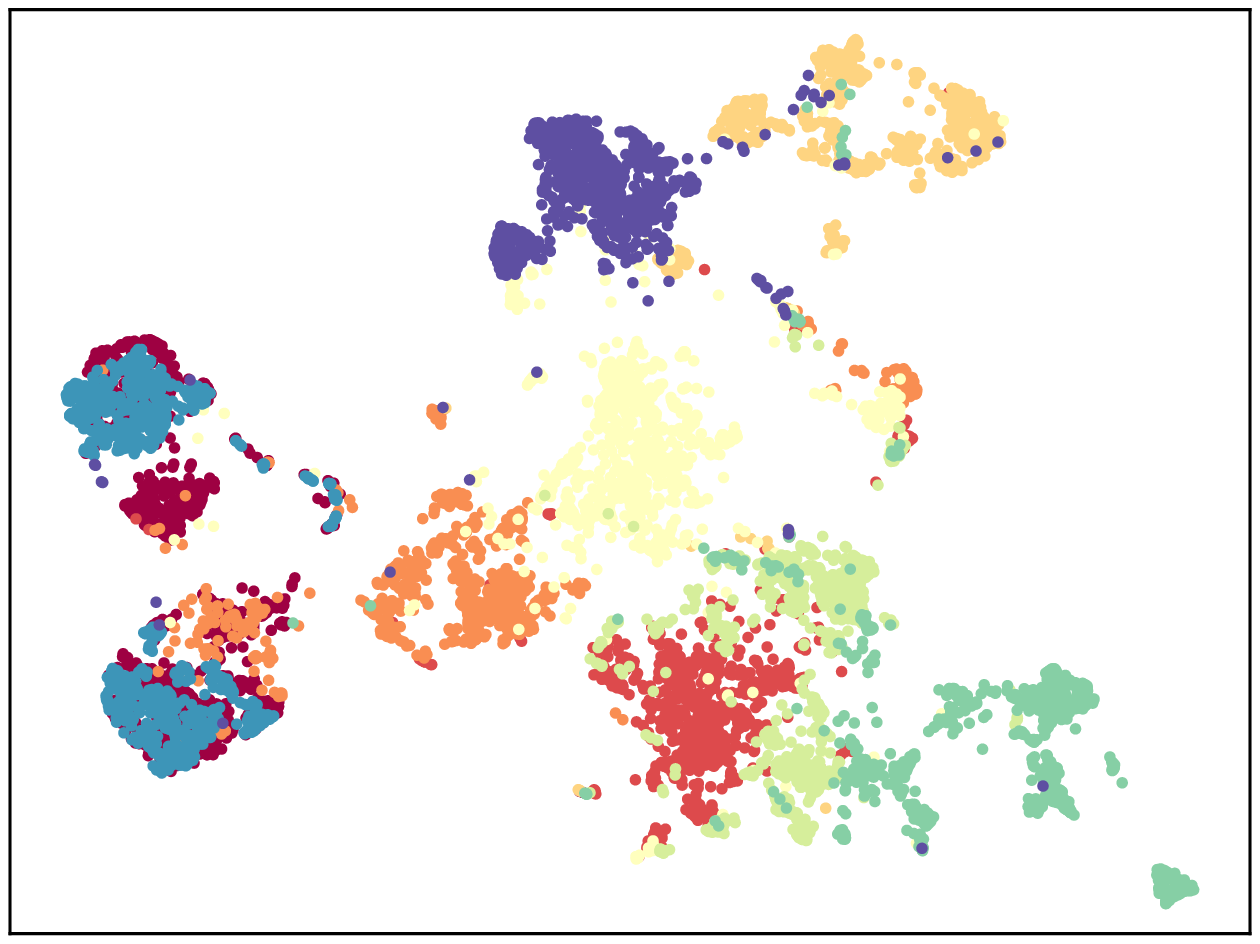}
			\end{minipage}
			\begin{minipage}[b]{.22\linewidth}
				\centering
				\includegraphics[scale=0.23]{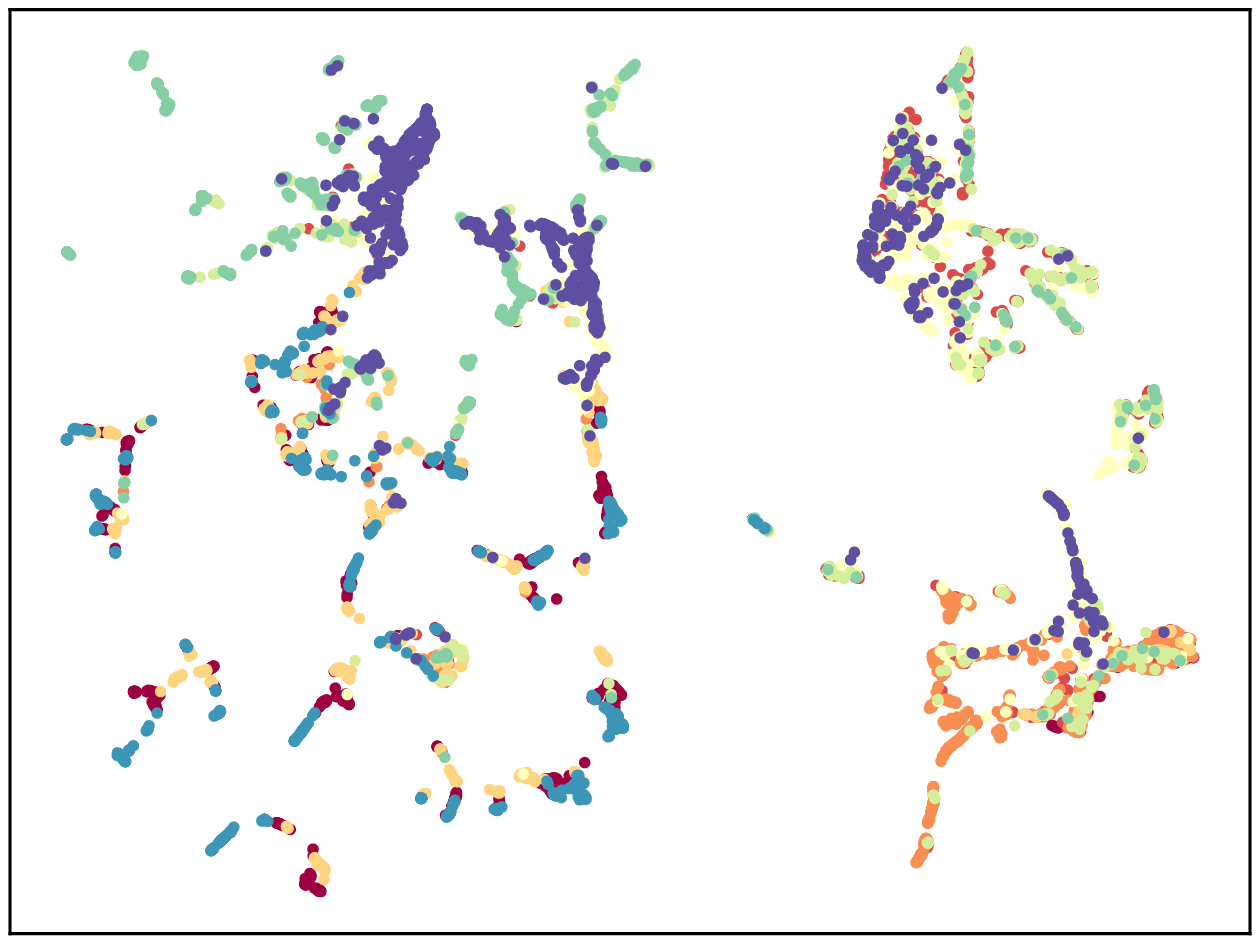}
			\end{minipage}
		}
		%\quad
		\subfigure[Training and validation curves]{
			\begin{minipage}[b]{.25\linewidth}
				\centering
				\includegraphics[scale=0.16]{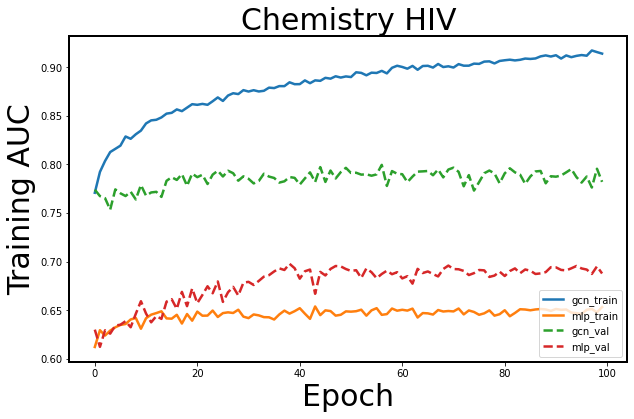}
			\end{minipage}
			\begin{minipage}[b]{.25\linewidth}
				\centering
				\includegraphics[scale=0.16]{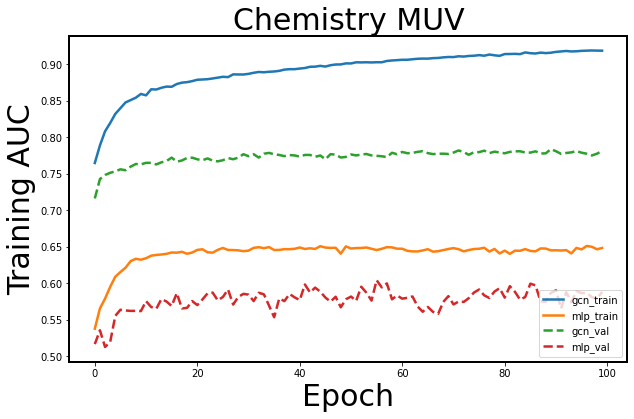}
			\end{minipage}
		}
		\caption{Experimental results of GCN and FCN. (a) GCN get better clustering quality and higher MGS value than FCN. (b)  Pre-trained GCN achieves higher training and validation ROC-AUC than pre-trained FCN on downstream classifaction task.}
		\label{fig2}
	\end{figure*}
	\subsection{Graph Fingerprint}
	
	Given a set of graphs$ \left\{G_{1},\ldots,G_{N}\right\} $, the graph fingerprint generation takes the high dimensional graph object and reduces the complexity down to a fixed length vector. 
	
	For molecules, the most common type of fingerprint is a series of binary digits (bits), and each bit position monitors the presence or absence of structural fragments. There are many different fingerprints available~\cite{landrum2013rdkit}, such as Topological fingerprints, MACCS keys, and Morgan fingerprints. 
	
	For proteins, let $ A_{i} $ be the adjacency matrices of graph $ G_{i} $, and let $  L_{i} = D_{i} - A_{i} $ be the laplacians of the graph, where $ D_{i} $ is  the corresponding diagonal matrices of degrees. We choose the top k eigenvalues $ \left\{\lambda_{0},\ldots,\lambda_{k}\right\} $ of the laplacians as fingerprint. 
	\subsection{Graph Similarity}
	In this section, we introduce structural similarity based on fingerprints, and cosine similarity based on pre-trained embeddings as graph similarity.
	\subsubsection{Structural Similarity}
	For molecules, based on the assumption that similar molecules have a lot of fragments in common, we can calculate similarity between two fingerprints as structural similarity. There are many metric algorithms such as Tanimoto, Dice, Cosine, Sokal, and Russel. We list Tanimoto algorithm here, in which $ f_{i} $ and $ f_{j} $ are fingerprints of graph $ i $ and $ j $:
	\begin{equation}
	\operatorname{Tani}\left(f_{i}, f_{j}\right)=\frac{f_{i} \cdot f_{j}}{\sum_{b} f_{i b}+\sum_{b} f_{j b}-f_{i} \cdot f_{j}}
	\end{equation}
	
	For proteins, we define structural similarity as the euclidean distance between two graphs' top k  eigenvalues: 
	\begin{equation}
	\operatorname{Dis\left(\lambda_{i}, \lambda_{j}\right)}=\sum_{p=0}^{k}\left(\lambda_{i p}-\lambda_{j p}\right)^{2}
	\end{equation}

	\subsubsection{Embedding Similarity}
	For both molecules and proteins, we obtain each graph embedding $ h_{i} $ for the input graph by passing it through the pre-trained GNN encoder. Then we choose the cosine similarity, which uses the cosine value of the angle between two feature vectors of graphs $ i $ and $ j $ to measure the similarity: 
	\begin{equation}
	\operatorname{Cos}(h_{i}, h_{j})=\frac{h_{i} \cdot h_{j}}{\|h_{i}\|\|h_{j}\|}
	\end{equation}
	
	\subsection{Correlation}
	After calculating the structural similarity and embedding similarity of a set of graph pairs, how to better characterize the relationship between the two kinds of similarity? A recent approach is from struc2vec~\cite{ribeiro2017struc2vec}, which proposes a novel and flexible framework to learn representations that capture the structural identity of nodes in a network. Struc2vec computes the spearman and pearson correlation coefficients between structural distance and the euclidean distance of all node pairs, and the experiment results corroborate that there is a very strong correlation between the two kinds of distances, which suggests that struc2vec indeed captures in the latent space the structural information adopted by the methodology. 
	
	Therefore, inspired by the approach in struc2vec and considering the two sets of similarity values do not follow a normal distribution, we choose spearman correlation instead of pearson correlation coefficients as Metric for Graph Structure (MGS):
	
	\begin{equation}
	\operatorname{MGS} =\frac{\sum_{i=1}^{N}\left(s_{i}-\bar{s}\right)\left(e_{i}-\bar{e}\right)}{\sqrt{\sum_{i=1}^{N}\left(s_{i}-\bar{s}\right)^{2}} \sqrt{\sum_{i=1}^{N}\left(e_{i}-\bar{e}\right)^{2}}}
	\end{equation}
	where $ s_{i} $ and $ e_{i} $ are the rank of structural similarity and embedding similarity respectively. 
	
	\section{Proposed Pre-training Strategies}
	
	In this section, we use two simple yet intuitive cases to examine whether our proposed metric MGS can measure the capability of pre-training strategies in capturing structural information. Then we propose a pre-training strategy based on the metric (PGM).
	
	\subsection{An Experimental Investigation}
	\subsubsection{Case 1: Performance on GCN and FCN with the Same Pre-training Strategy}
	We take Graph Convolutional Networks (GCN) \cite{kipf2017semi-supervised} and the simplest Fully Connected Neural Network (FCN)\cite{li2018deeper} as an example. From their layer-wise propagation rules, we can see that GCN can capture more structural information than FCN, because GCN has the graph convolution matrix $ \tilde{D}^{-\frac{1}{2}} \tilde{A} \tilde{D}^{-\frac{1}{2}} $ (Eq. (\ref{con:inventoryflow})) applied on the left of the feature matrix:
	\begin{equation}
	H^{(l+1)}=\sigma\left(H^{(l)} \Theta^{(l)}\right)
	\end{equation}
	\vspace{-0.4cm}
	\begin{equation}
	H^{(l+1)}=\sigma\left(\tilde{D}^{-\frac{1}{2}} \tilde{A} \tilde{D}^{-\frac{1}{2}} H^{(l)} \Theta^{(l)}\right)\label{con:inventoryflow}
	\end{equation}
	
	Therefore, we pre-train GCN and FCN in a supervised manner, finetune the pre-trained model to downstream classification task on HIV and MUV dataset from MoleculeNet~\cite{wu2018moleculenet}, a recently-curated benchmark for molecular property prediction, and then apply our metric and visualization method to pre-trained embedding. From Figure. \ref{fig2}, GCN shows excellent performance than FCN, because GCN extracts information from both the node features and the topological structures. From Figure. \ref{fig2}(a) GCN has better clustering effect, and higher MGS value. In addition, when finetune the pre-trained model to downstream classifaction task, pre-trained GCN achieves higher training and validation
	ROC-AUC than pre-trained FCN as shown in Figure. \ref{fig2}(b).
		\begin{figure*}[t]
		\begin{center}
			\centerline{\includegraphics[scale=0.48]{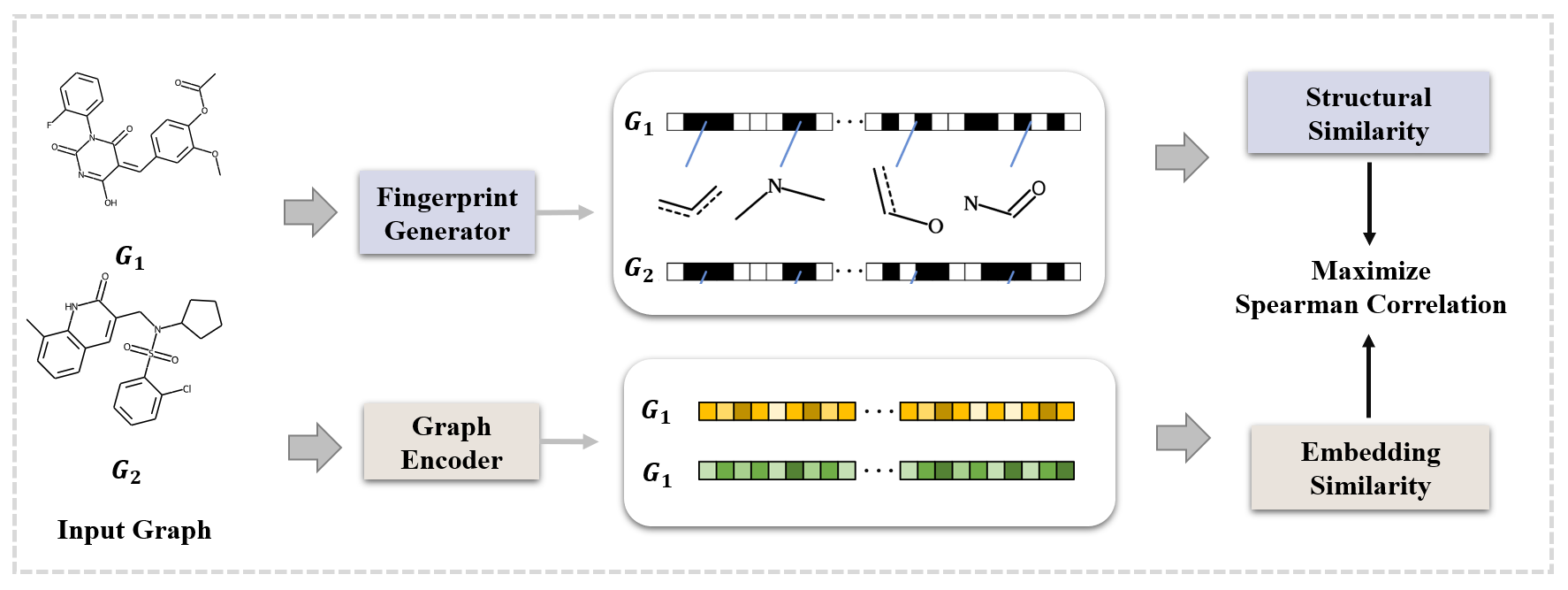}}
			\caption{The framework of PGM model. We first randomly sample a set of graph pairs, then generate graph fingerprint and encode graph embedding of each graph respectively. And then, structural similarity and embedding similarity of each pair is calculated. Finally, we maximize the Spearman correlation between the two kinds of similarity.}
			\label{fig3}
		\end{center}
	\end{figure*}
		\begin{table}
		\setlength{\tabcolsep}{1mm}
		\begin{center}
			{\caption{MGS Values On GIN with Different Pre-training Strategies.}
				\label{tab6}}
			\begin{tabular}{cccccc}
				\toprule
				&No-pretrain&Edgepred&Infomax&Contextpred&PGM\\
				\midrule
				MGS & 0.20&0.54&0.70&0.80&\textbf{0.98}\\
				\bottomrule
			\end{tabular}
		\end{center}
	\end{table} 
	  
	\subsubsection{Case 2: MGS Values on GIN with Different Pre-training Strategies} 
	We pre-train GIN using different pre-training strategies, which focus on capturing graph structure information. From Table \ref{tab6}, we can observe that MGS values on GIN with structural pre-training strategies is higher than that with no pretrain, and our proposed strategies PGM achieves the highest MGS value. 
	
	\subsection{Learning Objective}
	The above experiments demonstrate that the more structural information contained in graph embedding, the higher MGS value is. 
	Then an intuitive pre-training strategy is to optimize the metric as the loss shown in Figure~\ref{fig3}, that is, to maximize correlation between structural similarity and embedding similarity:
	\begin{equation}
	Loss = Correlation(Sim_{structural},Sim_{embedding})
	\end{equation}
	When pre-training is finished, we fine-tune the pre-trained GNN on downstream tasks. Specifically, we add linear classifiers on top of graph-level representations to predict downstream graph labels.

	\begin{table*}[t]
		\caption{Test ROC-AUC score on molecular and protein prediction using different pre-training strategies. (Left) Pre-train the GIN model on 2 million molecules sampled from the ZINC15, fine-tune on 8 downstream task for 10 times. (Right) Pre-train the GIN model on 395K protein ego-networks, fine-tune on 40 fine-grained prediction tasks.}
		\label{tab1}
		\resizebox{\textwidth}{!}{
			%\scalebox{0.69}{
			\begin{tabular}{cccccccccc|c}
				\toprule
				\textbf{Dataset} & \textbf{BBBP} & \textbf{Tox21} & \textbf{ToxCast} & \textbf{SIDER} & \textbf{ClinTox} & \textbf{MUV} & \textbf{HIV} & \textbf{BACE} & \textbf{Average} & \textbf{Protein}\\
				\midrule
				No.Molecules       & 2039  & 7831 & 8575 & 1427 & 1478 & 93087 & 41127 & 1513 & / &/\\
				No.Binary prediction tasks   & 1 & 12 & 617&  27 & 2 & 17 & 1 & 1 & / &/\\
				\midrule
				Pre-training strategy & \multicolumn{9}{c}{Average} & \\
				\midrule
				No Pre-train & 65.8 $\pm$ 4.5 & 74.0 $\pm$ 0.8 & 63.4 $\pm$ 0.6 & 57.3 $\pm$ 1.6 & 58.0 $\pm$ 4.4 & 71.8 $\pm$ 2.5 & 75.3 $\pm$ 1.9 & 70.1 $\pm$ 5.4 & 67.0 &64.8$\pm$1.0\\
				Infomax & 68.8 $\pm$ 0.8 & 75.3 $\pm$ 0.5 & 62.7 $\pm$ 0.4 & 58.4 $\pm$ 0.8 & 69.9 $\pm$ 3.0 & 75.3 $\pm$ 2.5 & 76.0 $\pm$ 0.7 & 75.9 $\pm$ 1.6 & 70.3 &64.1$\pm$1.5\\
				EdgePred & 
				67.3 $\pm$ 2.4 & 76.0 $\pm$ 0.6 & 64.1 $\pm$ 0.6 & 60.4 $\pm$ 0.7 & 64.1 $\pm$ 3.7 & 74.1 $\pm$ 2.1 & 76.3 $\pm$ 1.0 & 79.9 $\pm$ 0.9 & 70.3&65.7$\pm$1.3\\
				ContextPred & 68.0 $\pm$ 2.0 & 75.7 $\pm$ 0.7 & 63.9 $\pm$ 0.6 & 60.9 $\pm$ 0.6 & 65.9 $\pm$ 3.8 & \textbf{75.8 $\pm$ 1.7} & \textbf{77.3 $\pm$ 1.0} & 79.6 $\pm$ 1.2 & 70.9 &65.2$\pm$1.6 \\
				AttrMasking &64.3 $ \pm $ 2.8 &\textbf{76.7 $ \pm $ 0.4} & \textbf{64.2 $ \pm $ 0.5} & 61.0 $ \pm $ 0.7&71.8 $ \pm $ 4.1 & 74.7 $ \pm $ 1.4& 77.2 $ \pm $ 1.1  & 79.3 $ \pm $ 1.6 & 71.1 &64.4$\pm$1.3\\
				PGM&\textbf{69.4 $ \pm $ 1.5} &75.4 $ \pm $ 0.7 &63.7 $ \pm $ 0.5 &\textbf{61.8 $ \pm $ 0.7} & \textbf{74.8 $ \pm $ 2.7} &75.0 $ \pm $ 2.5 &74.3 $ \pm $ 1.0 & \textbf{81.7 $ \pm $ 1.3} & \textbf{72.0} &\textbf{67.5$\pm$1.1}\\
				\midrule 
				PGM-Infomax &\textbf{70.5 $ \pm $ 0.4} & 75.9 $ \pm $ 0.3 & 63.6 $ \pm $ 0.1 & \textbf{62.8 $ \pm $ 0.7} & 75.8 $ \pm $ 2.7 & 74.6 $ \pm $ 1.8 & 75.8 $ \pm $ 0.8 & 78.8 $ \pm $ 1.4 & 72.2 &66.6$\pm$1.4\\
				PGM-EdgePred &69.2 $ \pm $ 1.9 & 75.7 $ \pm $ 0.2 & 63.7 $ \pm $ 0.4 & 61.9 $ \pm $ 0.5 & 68.5 $ \pm $ 3.0 & 75.1 $ \pm $ 1.8 & 76.8 $ \pm $ 1.2 & 79.9 $ \pm $ 0.5 & 71.4 &\textbf{67.3$\pm$1.4}\\
				PGM-ContextPred & 69.1  $ \pm $ 2.0 &76.0 $ \pm $ 0.6 &63.5 $ \pm $ 0.3& 61.1 $ \pm $ 0.4 & 72.2 $ \pm $ 1.8 & 75.3 $ \pm $ 1.8 & \textbf{76.8 $ \pm $ 0.9} & 78.5 $ \pm $ 1.2 & 71.6 &66.5$\pm$1.4\\
				PGM-AttrMasking &68.2 $ \pm $ 1.8 &\textbf{76.1 $ \pm $ 0.3} & \textbf{63.8 $ \pm $ 0.2} & 60.4 $ \pm $ 0.8 & \textbf{77.4 $ \pm $ 2.2} & \textbf{77.2 $ \pm $ 1.3} & 75.3 $ \pm $ 0.9 & \textbf{81.2 $ \pm $ 1.1} & \textbf{72.4} &66.1$\pm$1.5\\
				\bottomrule
			\end{tabular}
		}
	\end{table*}
	\begin{table*}
		\caption{10-fold cross validation ROC-AUC score on molecular prediction benchmarks using different pre-training strategies. Pre-train the DeeperGCN model on the HIV dataset which contains 40K molecules, fine-tune on each downstream tasks.}
		\label{tab2}
		\resizebox{\textwidth}{!}{
			%\scalebox{0.69}{
			\begin{tabular}{ccccccccc}
				\toprule
				\textbf{Dataset} & \textbf{BBBP} & \textbf{Tox21} & \textbf{ToxCast} & \textbf{SIDER} & \textbf{ClinTox} & \textbf{HIV} & \textbf{BACE} & \textbf{Average}\\
				\midrule
				No Pre-train  & 82.1 $ \pm $ 1.7 & 76.1 $ \pm $ 0.6 & 63.3 $ \pm $ 0.8 & 55.7 $ \pm $ 1.4 & 75.0 $ \pm $ 3.6 & 73.4 $ \pm $ 0.9 & 72.8 $ \pm $ 2.1 & 71.2 \\
				\midrule
				
				InfoGraph & 80.4 $ \pm $ 1.2 & 76.1 $ \pm $ 1.1 & 64.4 $ \pm $ 0.8 & 56.9 $ \pm $ 1.8 & \textbf{78.4 $ \pm $ 4.0} & 72.6 $ \pm $ 1.0 & 76.1 $ \pm $ 1.6 & 72.1(+0.9 \%) \\
				
				\text { GPT-GNN } & 83.4 $ \pm $ 1.7 & 76.3 $ \pm $ 0.7 & 64.8 $ \pm $ 0.6 & 55.6 $ \pm $ 1.6 & 74.8 $ \pm $ 3.5 & 74.8 $ \pm $ 1.0 & 75.6 $ \pm $ 2.5 & 72.2(+1.0 \%) \\
				\text { GROVER } & 83.2 $ \pm $ 1.4 & 76.8 $ \pm $ 0.8 & 64.4 $ \pm $ 0.8 & 56.6 $ \pm $ 1.5 & 76.8 $ \pm $ 3.3 & 74.5 $ \pm $ 1.0 & 75.2 $ \pm $ 2.3 & 72.5(+1.3 \%) \\
				\text{GraphCL} & 82.8 $ \pm $ 1.0 & 76.1 $ \pm $ 1.3 & 63.3 $ \pm $ 0.5 &56.1 $ \pm $ 1.1&77.7 $ \pm $ 4.4 &75.0 $ \pm $ 1.0&76.5 $ \pm $ 2.8 & 72.5(+1.3\%)\\
				\text { MICRO-Graph } & 83.8 $ \pm $ 1.8 & 76.7 $ \pm $ 0.4 & 65.4 $ \pm $ 0.6 & 57.3 $ \pm $ 1.1 & 77.5 $ \pm $ 3.4 & 75.5 $ \pm $ 0.7 & 76.2 $ \pm $ 2.5 & 73.2(+2.0 \%) \\	
				\midrule
				
				PGM  & \textbf{89.0 $\pm$ 2.1} & \textbf{80.8 $\pm$ 2.1} & \textbf{70.2 $\pm$ 0.9} & \textbf{62.6 $\pm$ 2.6} & 73.3 $\pm$ 9.4 & \textbf{81.5 $\pm$ 1.8} & \textbf{85.5 $\pm$ 2.3} & \textbf{77.5(+8.8 \%)}\\
				
				\bottomrule
			\end{tabular}
		}
	\end{table*}
	\section{Experiment}
	
	In this section, we perform experiments on chemistry and biology domain following ~\cite{hu2020strategies}, which pre-train GNNs on graph data and fine-tune the pre-trained model on downstream tasks.
	
	\subsection{Datasets}
	For pre-training, we use 2 million unlabeled molecules sampled
	from the ZINC15 ~\cite{sterling2015zinc}  database for chemistry domain 
	and 395K unlabeled protein ego-networks derived
	from PPI networks of 50 species (e.g., humans, yeast, zebra fish) for biology domain.
	For downstream tasks, we use 8 binary classification datasets in MoleculeNet~\cite{wu2018moleculenet} for molecular property prediction and compose PPI networks from ~\cite{zitnik2019evolution}, consisting of 88K proteins from 8 different species for protein function prediction.
	
	\subsection{Experiment Setup}
	\subsubsection{Pre-training details} 
	We adopt
	a five-layer Graph Isomorphism Network (GIN) ~\cite{xu2019how} with 300-dimensional hidden units and a mean pooling readout function for performance comparisons. 
	We run all pre-training methods for 100 epochs, using Adam optimizer~\cite{kingma2015adam} (learning rate: 0.001) with a batch size of 256. 
	The Topological fingerprint and Tanimoto algorithm are selected to measure the structural similarity for molecules, and the top 6 eigenvalues of graph laplacian is set for protein fingerprint generation. 
	
	We compare the proposed pre-training strategy PGM with various competitive baseline strategies including 
	Deep Graph Infomax~\cite{velickovic2019deep}, EdgePred ~\cite{hamilton2017representation}, AttrMasking ~\cite{hu2020strategies}, ContextPred~\cite{hu2020strategies}, GROVER~\cite{rong2020self-supervised}, InfoGraph~\cite{sun2019infograph}, GPT-GNN~\cite{hu2020gpt-gnn}, GraphCL~\cite{you2020graph}, MICRO-Graph~\cite{zhang2020motif}.
	
	\subsubsection{Finetune details}
	On the molecular property prediction datasets,
	we train models for 100 epochs, while on the protein function prediction dataset (with the 40 binary prediction tasks), we train models for 50 epochs.
	We utilize grid search to find a good model configuration for some datasets (i.e. MUV in AttrMasking strategy). Unless otherwise specified, the batch size is set as 32, dropout rate is 0.5, and learning rate is 0.001.
	For evaluation, we select the test ROC-AUC at the best validation epoch. The
	downstream experiments are run with 10 random seeds, mean ROC-AUC and standard deviation is reported.

	\subsection{Results}
	\subsubsection{Performance on Downstream Tasks}
	We compare our method (PGM) with the previous strategies on classification datasets, which is shown in Table~\ref{tab1} and Table~\ref{tab2}. From the upper half of Table~\ref{tab1}, in the biology datasets, our PGM gains about 7.4$ \% $ performance enhancement against the no-pretrain baseline on average, and achieves comparable performance to the baseline strategies. Also, in the chemistry datasets, we see from the far right of Table~\ref{tab1} that our PGM achieves $ 4.2\% $ relative improvement
	over no-pretrain baseline. 
	From Table~\ref{tab2}, following the same experimental setting as MICRO-Graph~\cite{zhang2020motif}, we pre-trained DeeperGCN~\cite{li2020deepergcn} on the HIV dataset which contains 40K molecules.
	Then we fine-tuned the pre-trained model on downstream tasks. Table~\ref{tab2} shows
	that PGM outperforms previous best schemes on 6 of 7 datasets by a large margin (about 8.8\% average improvement).

	\subsubsection{Cooperation with Node-level Strategies} 
	From the lower half of Table~\ref{tab1}, we see that the performance of node-level strategies can be further enhanced when cooperating with our graph-level strategy. Furthermore, our method combining with AttrMasking outperforms that with other node-level strategies, achieving state-of-the-art performance on molecular prediction benchmarks. These results reveal that combinations of node-level and graph-level pre-trained strategies can capture more graph information.

	\subsubsection{Generalization over Different GNNs}
	We compare the pre-training gain with other popular GNN architectures (i.e. GCN~\cite{kipf2017semi-supervised}, GraphSAGE~\cite{hamilton2017inductive}, and GIN ~\cite{xu2019how}). We find GCN and GraphSAGE gains about 1.8$ \% $ and 2.5$ \% $ performance enhancement against the no-pretrain for chemistry domain, which reveals that our pre-training strategy achieves better performance than no pre-training baseline over different GNN architectures on most datasets. 
				\newcommand{\largefigsize}{0.08} % 右边那个图的大小
	\newcommand{\smallboxsize}{0.7} % 左边6张图整体的大小
	\newcommand{\smallfigsize}{0.2} % 每一张小图的大小
	\newcommand{\rheightsize}{0.7} % 右边图的相对纵坐标
	\newcommand{\vspacesize}{0} % 每个图片和名字之间的距离
	\newcommand{\hspacesize}{-3} % 整个图片和注释之间的距离
	\newcommand{\lspacesize}{3} % 整个图片和注释之间的距离
	
	\begin{figure*}[t] \centering 
		\raisebox{\rheightsize\height}{
			\makebox[\smallboxsize\textwidth]{
				\makecell{
					\includegraphics[width=\smallfigsize\textwidth]{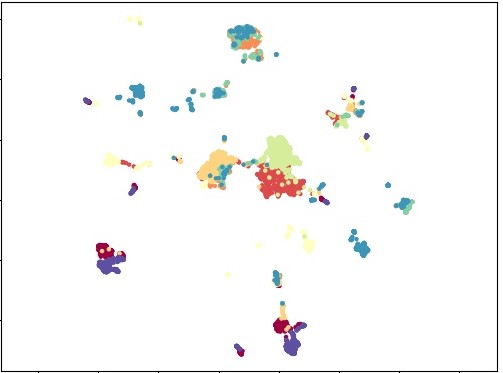}
					\hspace{\lspacesize pt}
					\includegraphics[width=\smallfigsize\textwidth]{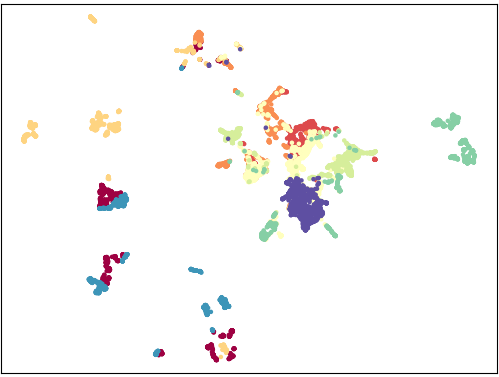}
					\hspace{\lspacesize pt}
					\includegraphics[width=\smallfigsize\textwidth]{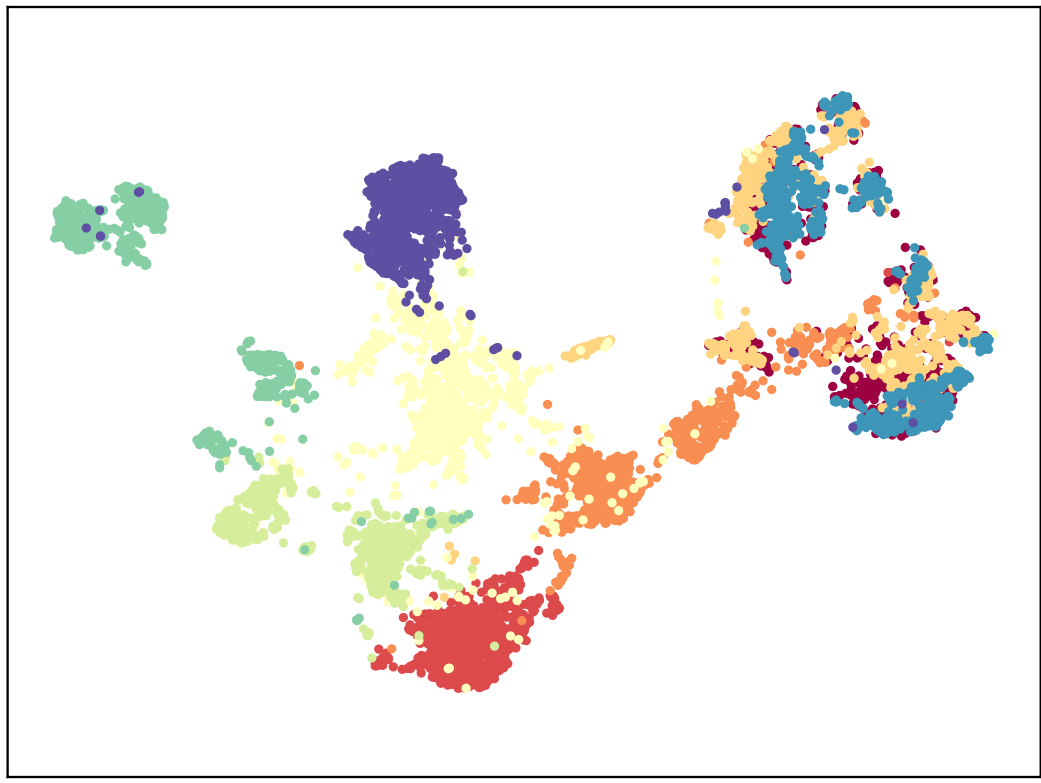}
					\hspace{\lspacesize pt}
					\\[\vspacesize pt]
					\makebox[0.20\textwidth]{\footnotesize (a) No-pretrain (0.20)}\qquad
					\makebox[0.20\textwidth]{\footnotesize (b) Edgepred (0.54)}\qquad
					\makebox[0.2\textwidth]{\footnotesize (c) Infomax (0.70)}
					\\
					\includegraphics[width=\smallfigsize\textwidth]{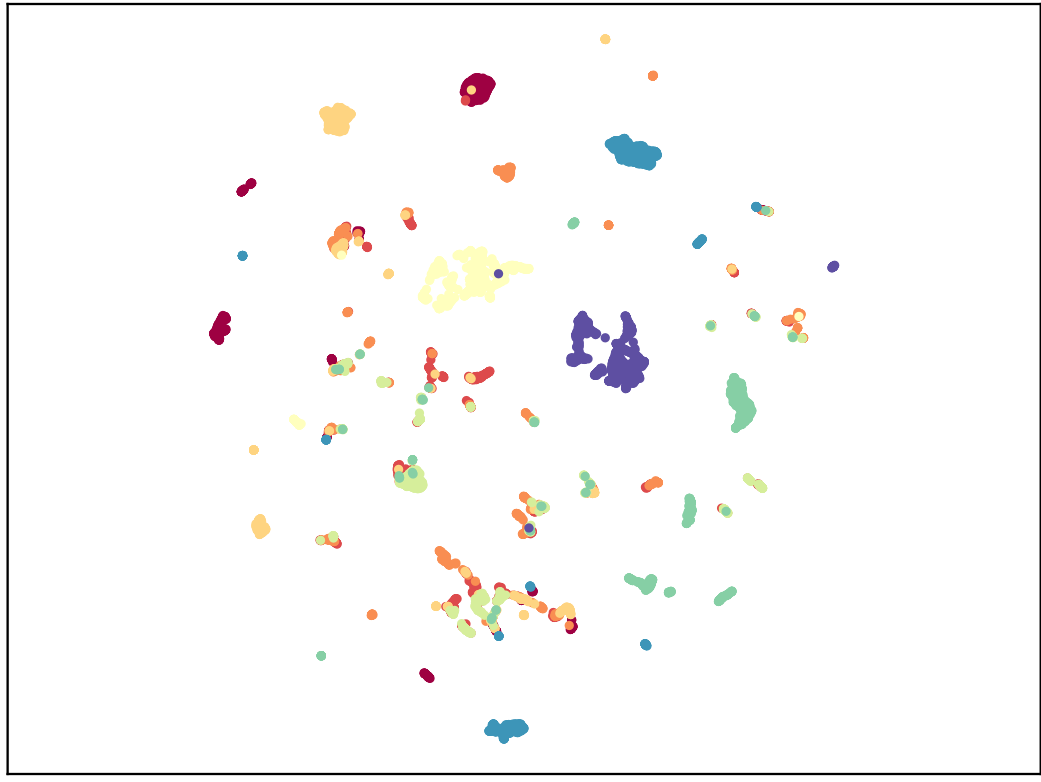}
					\hspace{\lspacesize pt}
					\includegraphics[width=\smallfigsize\textwidth]{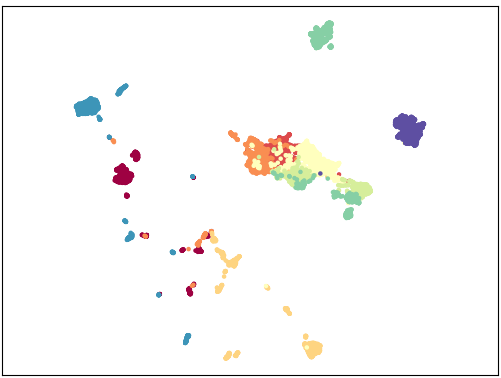}
					\hspace{\lspacesize pt}
					\includegraphics[width=\smallfigsize\textwidth]{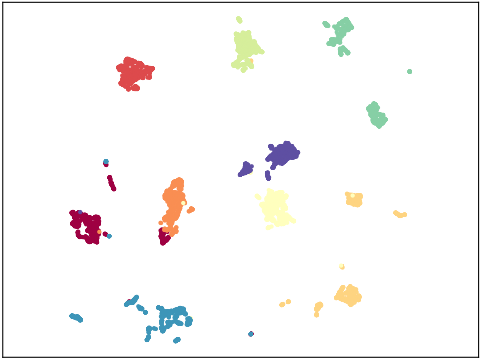}
					\\[\vspacesize pt]
					\makebox[0.2\textwidth]{\footnotesize (d) GROVER (0.73)}\qquad
					\makebox[0.2\textwidth]{\footnotesize (e) Contextpred (0.80)}\qquad
					\makebox[0.2\textwidth]{\footnotesize (f) PGM (0.98)}
				}
			}
		}
		\includegraphics[width=\largefigsize\textwidth]{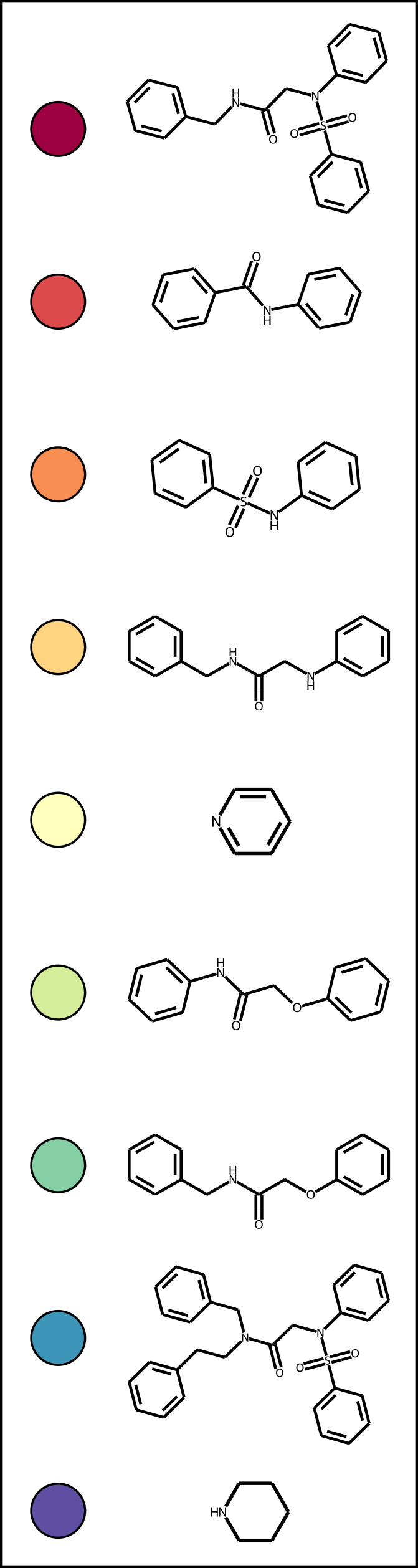}
		\hspace{\hspacesize pt}
		\\
		\caption{UMAP visualization of molecular graph representation coloring with the corresponding scaffold. The values in parentheses represent MGS metric. }
		\label{fig4}
	\end{figure*}
	\begin{table*}
		\begin{center}
			{\caption{Test ROC-AUC performance of different filters with and without pre-training.}
				\label{tab13}}
			\begin{tabular}{ccccccc}
				\toprule
				&\multicolumn{3}{c}{ Chemistry } & \multicolumn{3}{c}{ Biology } \\
				\midrule
				& Non-pretrained & Pre-trained &Gain & Non-pretrained & Pre-trained &Gain\\
				\midrule
				GIN & 67.0 &71.1 & $ +4.1 $ &64.8 $ \pm $ 1.0 & 64.4 $ \pm $ 1.3 & $ -0.4 $\\
				GCN & $68.9$ & $70.2$ &$ +1.3 $& $63.2 \pm 1.0$ & $61.5 \pm 1.3$ & $ -1.7 $\\
				ChebNet & \textbf{70.0} & \textbf{ 72.6 } &$ +2.6 $& $67.5 \pm 0.6$ & $68.7 \pm 0.5$ & $ +1.2 $ \\
				FAGCN & 67.0 & $68.1$ & $+1.1$ &\textbf{71.2 $\pm$ 0.8} & \textbf{71.6 $ \pm $ 0.4 } &$ +0.4 $\\
				\bottomrule
			\end{tabular}
		\end{center}
	\end{table*}

	\begin{table}
		\begin{center}
			{\caption{Test ROC-AUC average score on molecular prediction benchmarks with different fingerprint and similarity algorithms.}
				\label{tab3}}
			\begin{tabular}{ccc}
				\toprule
				\textbf{Fingerprint}& \textbf{Similarity} & \textbf{Average}\\
				\midrule
				\multicolumn{2}{c}{No Pre-train} & 67.0\\
				\midrule
				Morgan & Dice &70.6\\
				Morgan & Tanimoto & 70.4\\
				Topological & Dice & 70.3\\
				Topological & Tanimoto &\textbf{72.0}\\
				\bottomrule
			\end{tabular}
		\end{center}
	\end{table}

	\subsubsection{Performance with different Filter under Heterophily}
	We use Attrmasking as pre-training strategy to pre-train different filters. On the chemistry dataset, from Table \ref{tab13}, we observe that different filters all achieve improvement by pre-training. ChebNet performs best than other filters with and without pre-training, and achieves state-of-the-art performance in Tabel~\ref{tab1}. For biology domain, FAGCN performs best with and without pre-training, and achieves about 9.8\% relative improvment over GIN without pre-training. And FAGCN achieves about 5.5\% compared to the highest pre-trained method in Table \ref{tab1}. Therefore, based on the observation of the above results, we find that choosing the suitable filter sometimes may be better than designing a good pre-training strategy.

	\subsubsection{Other Fingerprint and Similarity Algorithms} In this experiment, we combine other fingerprint and similarity calculation algorithms, including Morgan fingerprint, Dice similarity algorithm. Morgan fingerprints, better known as circular fingerprints, is built by counting the number of all subgraphs within the setted radius starting from a specific atom. From the Table~\ref{tab3}, we can see that applying other molecular fingerprint and similarity algorithms to our PGM also achieves performance improvements than no pre-trained baseline. In addition, using Topological Fingerprints is better than Morgan Fingerprints, the reason may be that Topological Fingerprints starts by labelling molecules by storing atomic types, aromatic, and bond types in advance, which express the molecular structure more flexibly than Morgan Fingerprints. 

	\subsection{Visualization}
	For a more intuitive comparison of our proposed strategies' capability in capturing structural information, we adopt the same visualization procedure of PHD~\cite{li2021pairwise} on ZINC15 dataset. Specifically, we choose 9 most common scaffolds from ZINC15 dataset except benzene ring, and randomly sample 1000 molecules from each selected scaffold. Then the molecular graph representations of 9000 molecules are obtained by pre-trained models. Finally, we employed UMAP ~\cite{mcinnes2018umap} for molecular graph representation visualization coloring with the corresponding scaffold. 
	
	From Figure~\ref{fig4}, we observe that 
	the better clustering result there is, the higher the metric is. The no pre-trained GNN model (a) shows no obvious cluster pattern and the lowest MGS value, which means random initialization for model can not capture graph structure well. The graph representation learned from Infomax model (c) focus on local and global structure, which has more obvious cluster than Edgepred model (b). 
	The GROVER model (d) has many but clear clusters, the reason may be that the key idea of GROVER is to predict the class of functional groups, which makes pre-trained model capture local structure instead of global structure.
	The clustering result with our strategy (f) has a more compact structure, the highest intra-class similarity, and the clearest distinct boundaries, which reveals that our methods can better capture molecular global structural characteristics.
	
	\section{Conclusion}
	In this paper, we first verify the biological graph is heterophily, and explore the performance of low and high-pass filters in graph classification under heterophily. Experiments demonstrate that a appropriate filter sometimes may be better than designing a good pre-training strategy.
	Secondly, we design a quantitative metric (MGS), which utilizes the spearman correlation coefficient to measure the capability of pre-training strategies in capturing structural information. Through experimental investigations, we observe that there is a significantly high correlation between the metric and the model's capability in capturing graph structure. 
	Then, we propose a new self-supervised strategy PGM to pre-train an expressive graph neural network (GNN) at graph level to capture global structural pattern of graph. Extensive experiments show that our method is superior to many baseline algorithms. In the future, we plan to investigate the multi-task learning on graphs with node and graph-level pre-training strategies.

	\bibliography{mybib}

\begin{thebibliography}{10}

\bibitem{bo2021beyond}
Deyu Bo, Xiao Wang, Chuan Shi, and Huawei Shen, `Beyond low-frequency
  information in graph convolutional networks', in {\em Proceedings of the AAAI
  Conference on Artificial Intelligence}, volume~35, pp. 3950--3957, (2021).

\bibitem{defferrard2016convolutional}
Micha{\"e}l Defferrard, Xavier Bresson, and Pierre Vandergheynst,
  `Convolutional neural networks on graphs with fast localized spectral
  filtering', {\em Advances in neural information processing systems}, {\bf
  29}, (2016).

\bibitem{gilmer2017neural}
Justin Gilmer, S.~Samuel Schoenholz, F.~Patrick Riley, Oriol Vinyals, and
  E.~George Dahl, `Neural message passing for quantum chemistry', {\em In
  International Conference on Machine Learning (ICML)}, (2017).

\bibitem{hamilton2017inductive}
L.~William Hamilton, Rex Ying, and Jure Leskovec, `Inductive representation
  learning on large graphs', {\em In Advances in Neural Information Processing
  Systems (NeurIPS)},  1024--1034, (2017).

\bibitem{hamilton2017representation}
William~L Hamilton, Rex Ying, and Jure Leskovec, `Representation learning on
  graphs: Methods and applications', {\em arXiv preprint arXiv:1709.05584},
  (2017).

\bibitem{hu2020strategies}
Weihua Hu, Bowen Liu, Joseph Gomes, Marinka Zitnik, Percy Liang, Vijay Pande,
  and Jure Leskovec, `Strategies for pre-training graph neural networks', {\em
  arXiv preprint arXiv:1905.12265}, (2019).

\bibitem{hu2020gpt-gnn}
Ziniu Hu, Yuxiao Dong, Kuansan Wang, Kai-Wei Chang, and Yizhou Sun, `Gpt-gnn:
  Generative pre-training of graph neural networks', {\em In ACM SIGKDD
  Conference on Knowledge Discovery and Data Mining (KDD)},  1857--1867,
  (2020).

\bibitem{kingma2015adam}
Diederik~P Kingma and Jimmy Ba, `Adam: A method for stochastic optimization',
  {\em arXiv preprint arXiv:1412.6980}, (2014).

\bibitem{kipf2017semi-supervised}
Thomas~N Kipf and Max Welling, `Semi-supervised classification with graph
  convolutional networks', {\em arXiv preprint arXiv:1609.02907}, (2016).

\bibitem{landrum2013rdkit}
Greg Landrum, `Rdkit documentation', {\em Release}, {\bf 1}(1-79), ~4, (2013).

\bibitem{li2020deepergcn}
Guohao Li, Chenxin Xiong, Ali Thabet, and Bernard Ghanem, `Deepergcn: All you
  need to train deeper gcns', {\em arXiv preprint arXiv:2006.07739}, (2020).

\bibitem{li2021pairwise}
Pengyong Li, Jun Wang, Ziliang Li, Yixuan Qiao, Xianggen Liu, Fei Ma, Peng Gao,
  Sen Song, and Guotong Xie, `Pairwise half-graph discrimination - a simple
  graph-level self-supervised strategy for pre-training graph neural networks',
  {\em In International Joint Conference on Artificial Intelligence (IJCAI)},
  2694--2700, (2021).

\bibitem{li2018deeper}
Qimai Li, Zhichao Han, and Xiao-Ming Wu, `Deeper insights into graph
  convolutional networks for semi-supervised learning', pp. 3538--3545, (2018).

\bibitem{li2019label}
Qimai Li, Xiao-Ming Wu, Han Liu, Xiaotong Zhang, and Zhichao Guan, `Label
  efficient semi-supervised learning via graph filtering', in {\em Proceedings
  of the IEEE/CVF conference on computer vision and pattern recognition}, pp.
  9582--9591, (2019).

\bibitem{liu2021self}
Xiao Liu, Fanjin Zhang, Zhenyu Hou, Li~Mian, Zhaoyu Wang, Jing Zhang, and Jie
  Tang, `Self-supervised learning: Generative or contrastive', {\em IEEE
  Transactions on Knowledge and Data Engineering}, (2021).

\bibitem{mcinnes2018umap}
Leland McInnes, John Healy, and James Melville, `Umap: Uniform manifold
  approximation and projection for dimension reduction', {\em arXiv preprint
  arXiv:1802.03426}, (2018).

\bibitem{pei2020geom}
Hongbin Pei, Bingzhe Wei, Kevin Chen-Chuan Chang, Yu~Lei, and Bo~Yang,
  `Geom-gcn: Geometric graph convolutional networks', {\em arXiv preprint
  arXiv:2002.05287}, (2020).

\bibitem{ribeiro2017struc2vec}
Filipe Rodrigues~Leonardo Ribeiro, H.~P.~Pedro Saverese, and R.~Daniel
  Figueiredo, `struc2vec: Learning node representations from structural
  identity', {\em In ACM SIGKDD Conference on Knowledge Discovery and Data
  Mining (KDD)},  385--394, (2017).

\bibitem{rong2020self-supervised}
Yu~Rong, Yatao Bian, Tingyang Xu, Weiyang Xie, Ying Wei, Wenbing Huang, and
  Junzhou Huang, `Self-supervised graph transformer on large-scale molecular
  data', {\em In Advances in Neural Information Processing Systems (NeurIPS)},
  {\bf 33},  12559--12571, (2020).

\bibitem{sterling2015zinc}
Teague Sterling and J~John Irwin, `Zinc 15 - ligand discovery for everyone',
  {\em Journal of Chemical Information and Modeling},  2324--2337, (2015).

\bibitem{sun2019infograph}
Fan-Yun Sun, Jordan Hoffmann, Vikas Verma, and Jian Tang, `Infograph:
  Unsupervised and semi-supervised graph-level representation learning via
  mutual information maximization', {\em arXiv preprint arXiv:1908.01000},
  (2019).

\bibitem{velickovic2018graph}
Petar Veli{\v{c}}kovi{\'c}, Guillem Cucurull, Arantxa Casanova, Adriana Romero,
  Pietro Lio, and Yoshua Bengio, `Graph attention networks', {\em arXiv
  preprint arXiv:1710.10903}, (2017).

\bibitem{velickovic2019deep}
Petar Veli{\v{c}}kovi{\'c}, William Fedus, William~L Hamilton, Pietro Li{\`o},
  Yoshua Bengio, and R~Devon Hjelm, `Deep graph infomax', {\em arXiv preprint
  arXiv:1809.10341}, (2018).

\bibitem{wu2019simplifying}
Felix Wu, Amauri Souza, Tianyi Zhang, Christopher Fifty, Tao Yu, and Kilian
  Weinberger, `Simplifying graph convolutional networks', in {\em International
  conference on machine learning}, pp. 6861--6871. PMLR, (2019).

\bibitem{wu2018moleculenet}
Zhenqin Wu, Bharath Ramsundar, N.~Evan Feinberg, Joseph Gomes, Caleb Geniesse,
  S.~Aneesh Pappu, Karl Leswing, and S.~Vijay Pande, `Moleculenet: A benchmark
  for molecular machine learning', {\em Chemical science},  513--530, (2018).

\bibitem{xiao2020self-supervised}
Liu Xiao, Zhang Fanjin, Hou Zhenyu, Wang Zhaoyu, Mian Li, Zhang Jing, and Tang
  Jie, `Self-supervised learning: Generative or contrastive', {\em IEEE
  Transaction on Knowledge and Data Engineering}, (2020).

\bibitem{xu2019how}
Keyulu Xu, Weihua Hu, Jure Leskovec, and Stefanie Jegelka, `How powerful are
  graph neural networks?', {\em In International Conference on Learning
  Representations (ICLR)}, (2019).

\bibitem{xu2021self-supervised}
Minghao Xu, Hang Wang, Bingbing Ni, Hongyu Guo, and Jian Tang, `Self-supervised
  graph-level representation learning with local and global structure', {\em In
  International Conference on Machine Learning (ICML)},  11548--11558, (2021).

\bibitem{yan2022two}
Yujun Yan, Milad Hashemi, Kevin Swersky, Yaoqing Yang, and Danai Koutra, `Two
  sides of the same coin: Heterophily and oversmoothing in graph convolutional
  neural networks', in {\em 2022 IEEE International Conference on Data Mining
  (ICDM)}, pp. 1287--1292. IEEE, (2022).

\bibitem{you2020graph}
Yuning You, Tianlong Chen, Yongduo Sui, Ting Chen, Zhangyang Wang, and Yang
  Shen, `Graph contrastive learning with augmentations', {\em In Advances in
  Neural Information Processing Systems (NeurIPS)}, {\bf 33},  5812--5823,
  (2020).

\bibitem{zhang2020motif}
Shichang Zhang, Ziniu Hu, Arjun Subramonian, and Yizhou Sun, `Motif-driven
  contrastive learning of graph representations', {\em arXiv preprint
  arXiv:2012.12533}, (2020).

\bibitem{zhu2021graph}
Jiong Zhu, Ryan~A Rossi, Anup Rao, Tung Mai, Nedim Lipka, Nesreen~K Ahmed, and
  Danai Koutra, `Graph neural networks with heterophily', in {\em Proceedings
  of the AAAI Conference on Artificial Intelligence}, volume~35, pp.
  11168--11176, (2021).

\bibitem{zhu2020beyond}
Jiong Zhu, Yujun Yan, Lingxiao Zhao, Mark Heimann, Leman Akoglu, and Danai
  Koutra, `Beyond homophily in graph neural networks: Current limitations and
  effective designs', {\em Advances in Neural Information Processing Systems},
  {\bf 33},  7793--7804, (2020).

\bibitem{zitnik2019evolution}
Marinka Zitnik, Rok Sosi{\v{c}}, Marcus~W Feldman, and Jure Leskovec,
  `Evolution of resilience in protein interactomes across the tree of life',
  {\em Proceedings of the National Academy of Sciences}, {\bf 116}(10),
  4426--4433, (2019).

\end{thebibliography}
\end{document}